\documentclass{Interspeech}

\usepackage{subfigure}
\usepackage{amsthm,amsmath,amssymb}
\usepackage{mathrsfs}
\usepackage{multirow}
\usepackage{makecell}
\usepackage{enumitem}
\usepackage{stmaryrd}
\usepackage[table,xcdraw,dvipsnames]{xcolor}
\usepackage{listings}
\usepackage{color}
\usepackage{float}
\definecolor{dkgreen}{rgb}{0,0.6,0}
\definecolor{gray}{rgb}{0.5,0.5,0.5}
\definecolor{mauve}{rgb}{0.58,0,0.82}
\usepackage[ruled, vlined, linesnumbered]{algorithm2e}
\interspeechcameraready

\title{Novel Parasitic Dual-Scale Modeling for Efficient and Accurate Multilingual Speech Translation
\thanks{$^{\dagger}$Yanmin Qian is corresponding author}
}

\author[affiliation={1}]{Chenyang}{Le}
\author[affiliation={2}]{Yinfeng}{Xia}
\author[affiliation={2}]{Huiyan}{Li}
\author[affiliation={2}]{Manhong}{Wang}
\author[affiliation={2}]{Yutao}{Sun}
\author[affiliation={2}]{Xingyang}{Ma}
\author[affiliation={1}]{Yanmin}{Qian$^{\dagger}$}
\affiliation{}{}{$^1$Auditory Cognition and Computational Acoustics Lab}
\affiliation{}{}{MoE Key Lab of Artificial Intelligence, AI Institute}
\affiliation{}{}{School of Computer Science, Shanghai Jiao Tong University, Shanghai, China}
\affiliation{}{}{$^2$Honor Device Co, Ltd, China}
\email{\{nethermanpro,yanminqian\}@sjtu.edu.cn}


\usepackage{comment}

\begin{document}

\maketitle

\begin{abstract}
Recent advancements in speech-to-text translation have led to the development of multilingual models capable of handling multiple language pairs simultaneously. However, these unified models often suffer from large parameter sizes, making it challenging to balance inference efficiency and performance, particularly in local deployment scenarios. We propose an innovative Parasitic Dual-Scale Approach, which combines an enhanced speculative sampling method with model compression and knowledge distillation techniques. Building on the Whisper Medium model, we enhance it for multilingual speech translation into whisperM2M, and integrate our novel KVSPN module, achieving state-of-the-art (SOTA) performance across six popular languages with improved inference efficiency. KVSPN enables a 40\% speedup with no BLEU score degradation. Combined with distillation methods, it represents a 2.6$\times$ speedup over the original Whisper Medium with superior performance.
\end{abstract}

\section{Introduction}

In recent years, the field of speech-to-text translation has witnessed remarkable progress\cite{chenMAESTROMatchedSpeech2022, zhang2023Google, le2023ComSL, communication2023SeamlessM4T}, driven by the exponential growth of data availability and computational power. This advancement has led to the emergence of multilingual models capable of handling multiple language pairs simultaneously. Compared to training separate models for each language pair, a unified multilingual approach offers superior knowledge transfer and enhanced performance. However, existing multilingual translation models often suffer from large parameter sizes, making it challenging to balance inference efficiency and performance, particularly in scenarios requiring local deployment.

While various acceleration techniques have been proposed in the speech domain, their applicability to multilingual speech translation remains limited. Model distillation\cite{gandhi2023distil, Ferraz2024multilingual}, while effective for simpler tasks like speech recognition, faces constraints in multilingual settings due to the inherent complexity of understanding and generating multiple languages. For example, Whisper\cite{radford2023robust} has its official distilled turbo version, which pitifully no longer supports speech translation, for which a probable reason is that the distilled model no longer contains enough parameters to handle the multi-lingual translation task. Alternative approaches, such as speculative sampling\cite{leviathan2023fast, kim2024speculative} or multi-head Medusa mechanisms\cite{segal2024whisper}, prove less effective when applied to already compact models, as they often result in significant performance degradation. These limitations highlight the need for a more tailored approach to accelerate inference in multilingual speech translation tasks.

In this work, we present an innovative Parasitic Dual-Scale Approach, an enhanced speculative sampling method specifically designed for multilingual speech translation. Our approach combines model compression and knowledge distillation techniques commonly used in speech translation, achieving state-of-the-art (SOTA) performance while maintaining model compactness and efficiency. Building upon the Whisper Medium model, we first conduct task-specific fine-tuning for multilingual speech translation. Subsequently, we integrate our novel KVPSN (Key-Value Parasitic Speculative Network) module. Compared to previous methods, the KVPSN is more tightly integrated with the base model, which significantly improves inference efficiency with no performance degradation.

The contributions of this work are as follows:

\begin{enumerate}

\item We present whisperM2M, a modified version of Whisper, fine-tuned with extensive data, achieving SOTA performance in multilingual translation across six selected languages with reduced parameter size. Notably, our training data primarily consists of open-source resources.

\item We introduce the innovative KVPSN module, which achieves an additional 40\% inference speedup on top of an already efficient model, with no BLEU score reduction. This represents a 2.6$\times$ over the original Whisper Medium's inference speed.

\item As a speculative sampling approach, KVPSN maintains compatibility with other acceleration techniques (e.g., flash attention\cite{dao2022flashattention} or quantization\cite{shao2024dq}) and offers modular flexibility, allowing users to balance performance and efficiency according to their specific needs.

\end{enumerate}

\begin{figure*} %[t]

  \centering
  \includegraphics[width=0.95\textwidth]{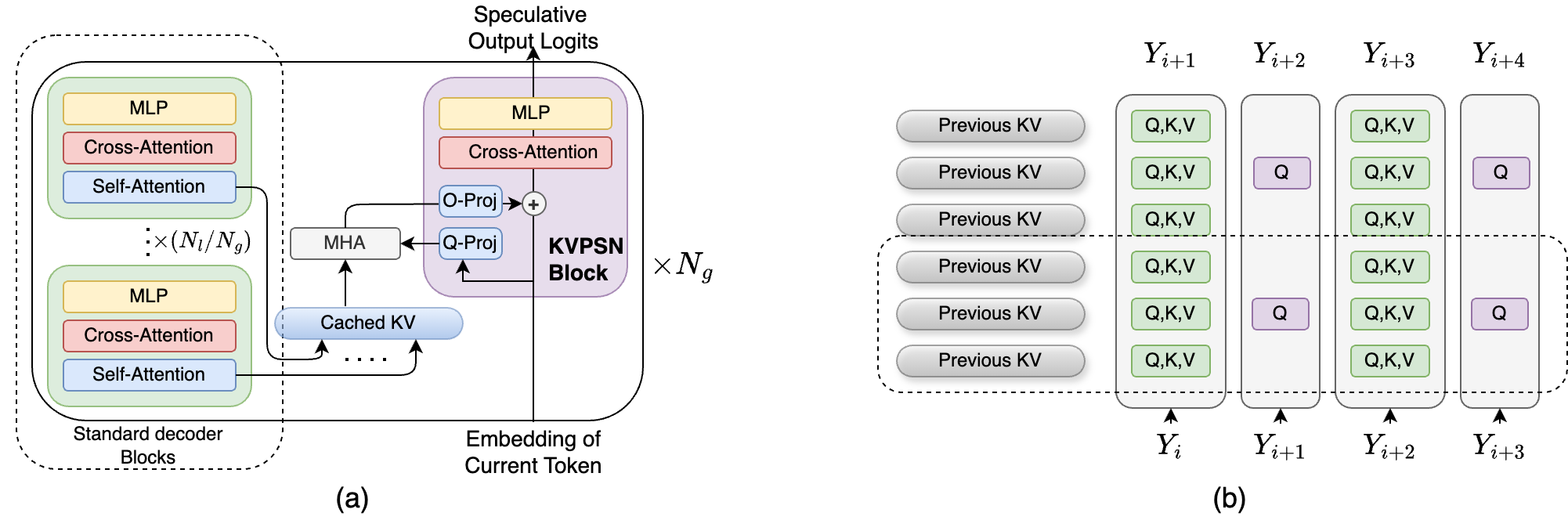}
  \caption{(a) Structure of one KVPSN block. (b) Example of inference process of KVPSN. Here the $N_l=6$ and $N_g=2$, thus the average inference cost reduces from 6 layers to 4 layers.}
\label{fig1}
\end{figure*}

\section{Methods}

\subsection{WhisperM2M}
\label{whisperm2m}

Whisper is a speech recognition system that follows the standard encoder-decoder transformer architecture. It is known for its multi-lingual recognition ability. To better serve the task of efficient speech translation, we make the following modification to the original whisper model into the whisperM2M model.

First, the language ID in the prompt of the original whisper model indicates the input language, which is unnecessary in translation tasks. In our reformulated version, we repurpose the language ID to indicate the language of the target output language.  We also removed other elements from the original prompt that are irrelevant to our task. 

Second, many previous works have shown promising results in training MT and ST tasks together in a multi-task way\cite{liuEndtoEndSpeechTranslation2019, rubenstein2023AudioPaLM, le2023ComSL}, and thus the ST can distill knowledge from MT through a distribution matching loss like KL divergence. Following this, we add a text encoder to the whisper model for the MT task.  Because whisper has never trained on text-to-text tasks, we do not add KL loss until the MT loss falls below the ST loss. 

Finally, to speed up during inference, we follow the method in \cite{gandhi2023distil} to reduce the number of layers of the decoder, since the size of the decoder has the most significant impact on inference speed.  We keep the first and last layers while evenly discarding internal layers to reach the target layer size. 

All of the above make up our base model for multilingual speech-to-text translation, achieving a balance between efficiency and performance. The training objective is the weighted combination of ST loss, MT loss, and KL loss.
\begin{equation}\label{eq1}
    L_{base} = w_{st} * L_{st} + w_{mt} * L_{mt} + w_{kl} * L_{kl}
\end{equation}

\subsection{KVPSN}
\label{predictor}

In this section, we present the innovative Key-Value Parasitic Speculative Network (KVPSN), a lightweight yet effective transformer-based architecture designed to accelerate inference with minimal computational overhead. Similar to the Medusa method, after the original decoder generates a token, the KVPSN module generates a speculative future token through a lightweight network, thus reducing the overall inference latency. 

The main idea behind KVPSN is to resolve two key weaknesses of standard Medusa implementations. First,  traditional Medusa heads operate in isolation, where each subsequent head must blindly hypothesize the outputs of its predecessors. This lack of coordinated decision-making creates errors when divergent predictions occur across heads. Second, conventional Medusa implementations derive predictions from single-layer hidden states of the base model. For moderately sized architectures (e.g., $<1$B parameters), this restricts their predictive accuracy due to limited contextual representation.

\subsubsection{Architectural Overview}
The structure of the KVPSN module is shown in Figure \ref{fig1} (a).  Instead of hidden states of the base model, KVPSN takes the embedding of the current generated token as the initial input. This design strictly adheres to the autoregressive generation paradigm, ensuring full forward dependency during speculative prediction and thereby enhancing output determinism.

Given a base decoder with $N_l$ layers, we partition them into stratified groups (in this work $N_g$=3). Each KVPSN block interacts with a dedicated layer group through a cross-model attention mechanism. For the $l-$th KVPSN block processing the $i-$th sequence position:
\begin{equation}
    \mathrm{q}_{\text{SN}}^{i, l} = \mathcal{W}_{\theta'}^Q \cdot \mathrm{h}_{in}^{i,l}
\end{equation}
\begin{equation}
\mathrm{h}_{attn}^{i,l} = \mathcal{W}_{\theta'}^O \cdot \mathrm{MHA}\left( \mathrm{q}_{\text{SN}}^{i, l}, \mathbf{k}_{base}^{<i, G_l}, \mathbf{v}_{base}^{<i, G_l} \right)
\end{equation}
where denotes concatenation across layers in group $G_l = \left[\frac{N_l}{N_g}\times (l-1), \frac{N_l}{N_g}\times l\right)$. The subsequent residual connection follows standard transformer practice:
\begin{equation}
\mathrm{h}_{out}^{i,l+1} = \mathrm{h}_{in}^{i,l} + \mathrm{LayerNorm}(\mathrm{h}_{attn}^{i,l})
\end{equation}

Under this architecture, the KVPSN shares the KV of the past sequence with the base decoder. Therefore, it leverages multi-layer contextual signals from the base model without additional computation, enabling a higher prediction accuracy.  The remaining components (cross-attention and feed-forward networks) remain identical to the standard transformer architecture. 

The loss of the KVPSN module $L_{spec}$ is similar to the base model described in equation \ref{eq1}, as this module can also perform text-to-text translation. The total loss is a weighted combination of base and KVPSN.
% \vspace*{-0.4\baselineskip}
\begin{equation}
    L_{total} = L_{base} + w_{spec} * L_{spec}
% \vspace*{-0.4\baselineskip}
\end{equation}

\begin{table*}[htb]
\centering
\caption{The BLEU score(\%) of different models on CoVoST2 and Fleurs testset, where all results are based on greedy search.  Only parameters involved in the speech-to-text translation task are calculated. N.L. denotes the number of layers in the text decoder. ALPT denotes the average generation latency per token. R.Speed is the decoder inference speed relative to the whisper medium model.}
\label{t1}
\begin{tabular}{llcclcccccclcc}
\bottomrule
 \multirow{2}{*}{Methods}  &\multirow{2}{*}{$|$}& \multirow{2}{*}{Param} &\multirow{2}{*}{$N.L.$}&\multirow{2}{*}{$|$}& \multicolumn{2}{c}{CoVoST2} &           \multirow{2}{*}{$|$}& \multicolumn{3}{c}{Fleurs}&\multirow{2}{*}{$|$}& \multicolumn{1}{c}{\multirow{2}{*}{ALPT}} & \multicolumn{1}{c}{\multirow{2}{*}{R.Speed}} \\                            &&  &&& X-EN& EN-X&           & X-EN&  EN-X& X-X&& \multicolumn{1}{l}{} & \multicolumn{1}{c}{}                        \\ \hline
  &&  &&& & &  &&  & &&  &\\[-0.9em]

  Whisper Small && 244M &12&& 22.4& -&           & -&  -& -&&  8.1& 202\%\\
  \quad + finetune&& - &-&& 28.9& 28.0&           & 19.9 &  24.9 & 17.8 &&  8.1& 202\%\\
 Whisper Medium && 769M &24&& 29.8&  -&           & -&  -& -&& 16.5& 100\%\\
 \quad+ finetune && - &-&& 36.1& 37.6&           & 26.0&  31.3& 23.9&& 16.5& 100\%\\
  SeamlessM4T Medium& & 821M &12& & 34.4& 35.9& & 25.6& 27.1& 16.1& & 10.2&162\%\\
   SeamlessM4T Large v2& & 1.5B &24& & 38.3& 40.8& & 29.8& 31.5& 19.6& & 31.1&53\%\\
 \hline
   &&  &&& & &  &&  & &&  &\\[-0.9em]
 whisperM2M && 561M &12&& 37.0& 38.9&           & 26.4 &  32.4 & 25.1 && 8.7& 189\%\\
 % \quad + KVPSN (Top-2)  && 605M &12+3&& 36.4& 38.5&           & 25.7 &  31.9 & 24.6&& 6.2& 267\%\\
  \quad + KVPSN (top-1)&& 605M &12+3&& 37.0& 38.8&           & 26.5&  32.4& 25.1&& 6.4& 259\%\\
 \bottomrule
\end{tabular}
\end{table*}

\subsubsection{Speculative Execution}
\label{sec:dec}
Given the computational constraints imposed by the base model's scale, our implementation adopts a conservative single-token speculation strategy. As depicted in Figure \ref{fig1}(b), the inference process follows an interleaved execution pattern:

\begin{enumerate}
\item \textbf{Base Model Step:} Generates token $y_t$ with full autoregressive computation
\item (\textbf{Optional) Validation Step}: If $y_{t-1}$ is predicted by KVPSN, the base model validates it using its own output distribution. If denial, discard $y_t$ and replace $y_{t-1}$ with the output of the base model. In following steps, $t\xleftarrow{} t-1$.
\item \textbf{KVPSN Step}: Utilizes $y_t$'s embedding and the base model's KV cache to speculatively predict $y_{t+1}$.

\end{enumerate}

This tight coupling ensures computational state sharing – the KVPSN's speculative generation directly inherits the base model's contextual representation through shared KV cache access, eliminating redundant computation.

A top-k validation mechanism is implemented to control quality and mitigate error propagation in speculative decoding.
\vspace*{-0.8\baselineskip} 
\begin{equation}
\mathrm{Accept}(y_{t}) = \begin{cases}
1 & \text{if } y_{t} \in \mathrm{top}\text{-}k\left(P_{base}(y_{t}|y_{< t})\right) \\
0 & \text{otherwise}
\end{cases}
\vspace*{-0.8\baselineskip} 
\end{equation}
Here we use a fixed K. Thus the quality of speculative prediction will have a direct impact on inference efficiency.

\section{Experiments}

\subsection{Task \& Data}
We focus on many-to-many speech-to-text translation across six popular languages: EN, ZH, DE, ES, FR, and IT. We collect various open-source speech recognition datasets covering the six languages, including  LibriSpeech~\cite{panayotov2015librispeech}, Multilingual Librispeech(MLS) ~\cite{pratap2020mls}, VoxPopuli ~\cite{wang2021voxpopuli}, Common Voice\cite{ardila2019common}, WenetSpeech\cite{zhang2022wenetspeech}, KeSpeech\cite{tang2021kespeech} and Emilia~\cite{emilia}. Then we create pseudo translation labels for these data by translating the transcription into different languages through a cloud text-to-text translation API. The dataset consists of approximately 100K hours of training data, with about 80\% of it being in Chinese and English.

\subsection{Model}

As mentioned in section \ref{whisperm2m}, the whisperM2M model is initialized by the whisper medium model. A 6-layer text encoder, with a hidden size of 1024, is incorporated to facilitate the text-to-text translation task. Starting with a 24-layer encoder and 24-layer decoder whisper medium model, we add a 6-layer text encoder and prune the text decoder to $N_l = 12$ layers for decoding efficiency. Additionally, we perform an ablation study to compare the performance of the proposed model with an alternative 8-layer version.

The KVPSN module consists of $N_g = 3$ blocks, with the hidden size matching that of the base model. This leads to a theoretical maximum of $\frac{N_l-N_g}{N_l + N_g}=60\%$ computation speeding. And K=1 means to rollback every time the speculation deviates from the base model.

\subsection{Training}

To retain the capability of the whisper encoder,  we employ Low-Rank Adaptation(LoRA)\cite{hu2021lora}  ($\alpha = 64$ ) at the first 20 layers of the whisper encoder. The rest 4 encoder layers, as well as the text encoder, text decoder, and KVPSN module, are set free. The text encoder is initialized by the weights in self-attention and MLP modules of the first six whisper decoder layers. The KVPSN module is randomly initialized. For training loss, we set $w_{st}=2$, $w_{mt}=1$, $w_{kl}=0.5$ . Empirical study shows the model is not sensitive to these weights. And we set $w_{spec}=0.2$ to prioritize the training of the base model. It turns out that KVPSN converges well with this small weight. For each experiment train in fp16 on 8 Nvidia H800 GPUs for 2 million steps, using a batch size of 64 utterances. We use AdamW optimizer\cite{loshchilov2017decoupled} and linear learning rate schedular with 5000 steps of warmup and a maximum learning rate of 1e-4.

\subsection{Evaluation}

We report the case-sensitive BLEU\cite{papineni2002Bleu} with HF sacreBLEU implementation on two testset: CoVoST2\cite{wang2021covost} and Fleurs\cite{conneau2023fleurs}. We report the average BLEU score of 5 non-English to English pairs(X-EN) and the average score of 2 English to Non-English pairs(EN-X) provided by CoVoST2. Fleurs testset provides parallel data within our six languages, allowing many-to-many evaluation between all 30 pairs. We report the average score of 5 X-EN pairs, 5 EN-X pairs, and 30 X-X pairs.

As baselines, we compare our whisperM2M and KVPSN models against whisper small, whisper medium, Seamless M4t medium, and Seamless large v2\cite{communication2023Seamless}. For fairness, we fine-tune the two whisper models on our full training set, as they are not dedicately designed for translation.

For inference speed evaluation, we measure the average latency per token in milliseconds (ALTP). ALTP is computed as the total time for decoder generation divided by the total number of tokens generated. We sample 1000 utterances from the CoVoST2 test set for the inference speed test, running the process for 10 rounds and reporting the average ALTP. All evaluations are conducted on a single Nvidia H800 GPU using the Ubuntu platform.

\section{Results}

\subsection{Main Result}

Table~\ref{t1} shows the BLEU score and inference efficiency for greedy decoding on the testsets. The R.Speed is the relative inference speed against whisper medium models. When calculating the parameter size, only the parameters involved in speech-to-text translation are considered. When calculating ALTP we only consider the time in decoder auto-regressive generation and exclude the encoder forward. 

We can see from the table, that whisperM2M shows superior performance in speech translation than all other models except SeamlessM4T Large v2. It is noteworthy that whisperM2M even performs better than whisper medium that is finetuned on the same training set. This shows the following empirical results:  1) With proper training data, a whisper is a suitable base model for multi-lingual speech translation, as all of its variants perform well after finetuning  2)  Pruning layer is effective in improving efficiency in this task as 12 layers are enough to handle translation between six languages. 3) Knowledge distillation from the text-to-text translation task is helpful even if the training data is adequate. 

Compared to seamless models, WhisperM2M outperforms M4T medium model in all test sets. Although the large v2 model is powerful in English-centric translation, it marginally falls behind whisperM2M by 5.5 bleu score in Fleurs many-to-many settings. Possibly it is because of some issue in the distribution of training data while ours is more balanced. 

As for the efficiency, we can conclude from the table that within the same architecture, the inference speed is very related to the number of layers in the decoder. Thus the efficiency of whisperM2M is comparable to whisper small with the same number of decoder layers.  The KVPSN module takes a step further, increasing the inference speed of whisperM2M by 37\% to become the fastest model in the table with an extra 44M parameter. During this acceleration no performance is lost, thanks to the top1 validation rollback.  With this module, whisperM2M can reach a performance comparable to that of M4T large v2 with about 5 times the inference speed.

\subsection{Ablation Study}

\subsubsection{Decoding Configuration Analysis}
This section investigates the performance impacts of different speculative decoding configurations. We systematically adjust the threshold $k$ in our top-$k$ rollback validation mechanism (introduced in Section \ref{sec:dec}) and examine the integration with beam search algorithms. Evaluation is conducted on the CoVoST2 testset, reporting average BLEU scores across 7 language pairs.

Table \ref{tab:2} reveals three key observations for greedy decoding: \textbf{1.} Without rollback validation (maximum $k$), we achieve 48\% speedup – approaching the theoretical maximum of 60\% – but at the cost of a 1.2 BLEU score degradation \textbf{2.} Progressively reducing $k$ improves translation quality, with performance converging to baseline levels at $k=1$ (the minor BLEU difference stems from implementation-specific factors) \textbf{3.} The optimized configuration ($k=1$) maintains 37\% acceleration while preserving translation quality.

For beam search experiments ($N_b$=3), we observe distinct behavior: \textbf{1.} Rollback thresholds show minimal impact ($\leq$0.34 BLEU difference even without validation) \textbf{2.} KVPSN-enhanced beam search slightly outperforms baseline greedy decoding in both quality (+0.2 BLEU) and speed (+12\%).

This comparative analysis demonstrates that our method achieves different optimal operating points for greedy and beam search paradigms, providing flexibility for quality-speed tradeoffs.

\begin{table}[h]
    \centering
    \caption{Comparison of BLEU scores and speed for different decoding configurations.}
    \resizebox{0.47\textwidth}{!}{
    \begin{tabular}{llccccc}
    \toprule
        && Base & \multicolumn{4}{c}{w/ KVPSN} \\ 
        &&  - & top-1 & top-2 & top-3 & top-$\infty$ \\ 
        \hline
                &&  &  &  &&  \\[-0.9em]
        \multirow{2}{*}{Greedy} & BLEU & 37.51 & 37.55 & 37.03 & 36.68 & 36.28 \\
        & speed & 100\% & 137\% & 141\% & 143\% & 148\% \\ 
        \hline
                &&  &  &  &&  \\[-0.9em]
        \multirow{2}{*}{BS 3} & BLEU & 38.05 & 37.89 & 37.74 & 37.76 & 37.71 \\ 
        & speed & 84\% & 110\% & 113\% & 113\% & 116\% \\ 
    \bottomrule
    \end{tabular}
    }
    \label{tab:2}
\end{table}

% \begin{table}[h]
%     \centering
%     \caption{Comparison of BLEU scores and speed for different decoding configurations.}
%     \resizebox{0.47\textwidth}{!}{
%     \begin{tabular}{llcccc}
%     \toprule
%         & & top-1 & top-2 & top-3 & top-$\infty$ \\ 
%         \hline
%                 &  &  &  &&  \\[-0.9em]
%         \multirow{2}{*}{1 token} & BLEU  & 37.55 & 37.03 & 36.68 & 36.28 \\
%         & speed & 137\% & 141\% & 143\% & 148\% \\ 
%         \hline
%                 &  &  &  &&  \\[-0.9em]
%         \multirow{2}{*}{3 tokens} & BLEU  & 37.16 & - & - & 33.42 \\ 
%         & speed & 157\% & - & - & 203\% \\ 
%     \bottomrule
%     \end{tabular}
%     }
%     \label{tab:2}
% \end{table}

\subsubsection{Compare to different methods}

In this section, we compare KVPSN with two other techniques for boosting efficiency: 1) Further prune the decoder to 8 layers. This should result in a similar inference speech with KVPSN. 2) Medusa speculative decoding. For fairness, we use the same 3 transformer blocks for the additional Medusa heads. The result is shown in Table \ref{tab:3}.

For the pruning method, after further pruned to 8 layers, the model shows 2.6 bleu score degradation compared to the 12-layer base model. So 8 layers of decoder may not contain enough parameters to support this task. 

For Medusa methods, this table shows that the additional Medusa head does not perform well under the same training and inference setting. Without rollback, the Bleu score drops by 4.96. Even with top-1 validation, the acceleration is only 9\% due to too many rollbacks.

\begin{table}[]
\centering
\caption{Comparison of BLEU score and decoder inference speech between KVPSN and two other common techniques: further pruning and Medusa.Rel.S. is the relative speech to the base model with a 12-layer decoder. }
\label{tab:3}
\begin{tabular}{llccc}
\bottomrule
 Methods &Rollback& BLEU &ALTP& \multicolumn{1}{c}{Rel.S.} \\ \hline
Base&-& 37.51 &8.71 &100\%\\
Prune L8&-& 34.91 &6.02 &145\%\\
\hline
 \multirow{3}{*}{Medusa}&top-$\infty$& 32.55 &5.86&149\%\\
 & top-2& 36.54 &7.33&119\%\\
 & top-1&  37.50&8.14&107\%\\
 \hline
 \multirow{3}{*}{KVPSN}& top-$\infty$& 36.28 &5.87 &148\%\\
 
 &top-2& 37.03 &6.18& 141\%\\
 & top-1& 37.55  &6.36&137\%\\
 \bottomrule
\end{tabular}
\end{table}

\section{Conclusions and Discussions}

In this paper, we present the full workflow to transform whisper into a powerful and efficient model for multi-lingual speech translation. We present whisperM2M that reaches SOTA performance among models of similar size and efficiency. KVPSN module can boost efficiency by 37\% with no performance degradation.  

One thing to be discussed is that we choose to directly flatten the KV of different layers and allow queries from KVPSN to attend to each of them. This may cause a high computation cost if the sequence to be generated is too long($>$500 tokens). For generating longer sequences, it is recommended to combine with some KV compression techniques for better efficiency.

Currently, we only test our KVPSN method on a single task with encoder-decoder transformer. We hope this technique can be applied to more tasks and models. The architecture of KVPSN should be more coherent if applied to a decoder-only model with no cross-attention module. Moreover, we will try multi-token generation in speculation decoding, which may further increase efficiency.

\ifinterspeechfinal
    \section{Acknowledgement}
    This work was supported in part by China NSFC projects under Grants 62122050 and 62071288, in part by Shanghai Municipal Science and Technology Commission Project under Grant 2021SHZDZX0102.
\fi

% \clearpage
\bibliographystyle{IEEEtran}
\bibliography{mybib}

\end{document}